\pdfoutput=1

\documentclass[11pt]{article}

\usepackage[final]{acl}

\usepackage{amssymb}
\usepackage{pifont}
\usepackage{amsmath}

\usepackage{booktabs}
\usepackage{pdfpages}

\usepackage{graphicx}

\newcommand{\textred}[1]{\textcolor{red}{\textbf{#1}}}
\newcommand{\textgreen}[1]{\textcolor{green}{\textbf{#1}}}
\newcommand{\cmarkb}{\textgreen{\ding{52}}}
\newcommand{\xmark}{\textred{\ding{55}}}
\usepackage{latexsym}
\usepackage{enumitem}
\setenumerate[1]{itemsep=0pt,partopsep=0pt,parsep=\parskip,topsep=5pt}
\setitemize[1]{itemsep=0pt,partopsep=0pt,parsep=\parskip,topsep=5pt}
\setdescription{itemsep=0pt,partopsep=0pt,parsep=\parskip,topsep=5pt}

\usepackage{times}
\usepackage{latexsym}

\usepackage[T1]{fontenc}

\usepackage[utf8]{inputenc}

\usepackage{cleveref}
\crefname{section}{§}{§§}
\Crefname{section}{§}{§§}

\usepackage{microtype}

\usepackage{inconsolata}
\usepackage{subfigure}
\usepackage{multirow}
\usepackage{makecell}
\usepackage{balance}
\usepackage{listings}

\colorlet{punct}{red!60!black}
\definecolor{background}{HTML}{EEEEEE}
\definecolor{delim}{RGB}{20,105,176}
\colorlet{numb}{magenta!60!black}

\lstdefinelanguage{json}{
    basicstyle=\normalsize\ttfamily,
    numbers=left,
    numberstyle=\scriptsize,
    stepnumber=1,
    numbersep=5pt,
    showstringspaces=false,
    breaklines=true,
    frame=lines,
    backgroundcolor=\color{background},
    literate=
     *{0}{{{\color{numb}0}}}{1}
      {1}{{{\color{numb}1}}}{1}
      {2}{{{\color{numb}2}}}{1}
      {3}{{{\color{numb}3}}}{1}
      {4}{{{\color{numb}4}}}{1}
      {5}{{{\color{numb}5}}}{1}
      {6}{{{\color{numb}6}}}{1}
      {7}{{{\color{numb}7}}}{1}
      {8}{{{\color{numb}8}}}{1}
      {9}{{{\color{numb}9}}}{1}
      {:}{{{\color{punct}{:}}}}{1}
      {,}{{{\color{punct}{,}}}}{1}
      {\{}{{{\color{delim}{\{}}}}{1}
      {\}}{{{\color{delim}{\}}}}}{1}
      {[}{{{\color{delim}{[}}}}{1}
      {]}{{{\color{delim}{]}}}}{1},
}

\usepackage{soul}
\usepackage{color, xcolor}
\definecolor{cRed}{HTML}{DC9697}
\definecolor{cBlue}{HTML}{A7C5DC}
\definecolor{cYellow}{HTML}{F2E08A}
\definecolor{cGreen}{HTML}{ABD39E}

\usepackage{url}

\newcommand{\taskname}{{CogBench}}
\newcommand{\agent}{{CogGPT}}

\title{\agent: Unleashing the Power of Cognitive Dynamics on \\ Large Language Models}

\author{Yaojia Lv$^1$, Haojie Pan$^2$, Zekun Wang$^1$, Jiafeng Liang$^1$, Yuanxing Liu$^1$ \\
    \textbf{Ruiji Fu$^{2}$, Ming Liu$^1$, Zhongyuan Wang$^2$, Bing Qin$^1$} \\
    $^1$ Harbin Institute of Technology $^2$ Kuaishou Inc. \\
    \texttt{\{yjlv, zkwang, jfliang, yxliu, mliu, qinb\}@ir.hit.edu.cn} \\
    \texttt{\{panhaojie,furuiji,wangzhongyuan\}@kuaishou.com}
}

\begin{document}
\maketitle
\begin{abstract}

Cognitive dynamics, which refer to the evolution in human cognitive processes, are pivotal to advance human understanding of the world. Recent advancements in large language models (LLMs) highlight their potential for cognitive simulation. However, these LLM-based cognitive studies primarily focus on replicating human cognition in specific contexts, overlooking the inherently dynamic nature of cognition. To bridge this gap, we explore the cognitive dynamics of LLMs and present a corresponding task inspired by longitudinal studies. Toward the task, we develop \taskname, a novel benchmark to assess the cognitive dynamics of LLMs and validate it through participant surveys. We also design two evaluation metrics for \taskname, including Authenticity and Rationality. Recognizing the inherent static nature of LLMs, we further introduce \agent~for the task, which features an innovative iterative cognitive mechanism to develop lifelong cognitive dynamics. Empirical results demonstrate the superiority of \agent~over several existing methods, particularly in its ability to facilitate role-specific cognitive dynamics under continuous information flows. \footnote{Code and data are available at \url{https://github.com/KwaiKEG/CogGPT}}

\end{abstract}

\section{Introduction}
\begin{figure}[t]
  \centering 
  \includegraphics[width=\linewidth]{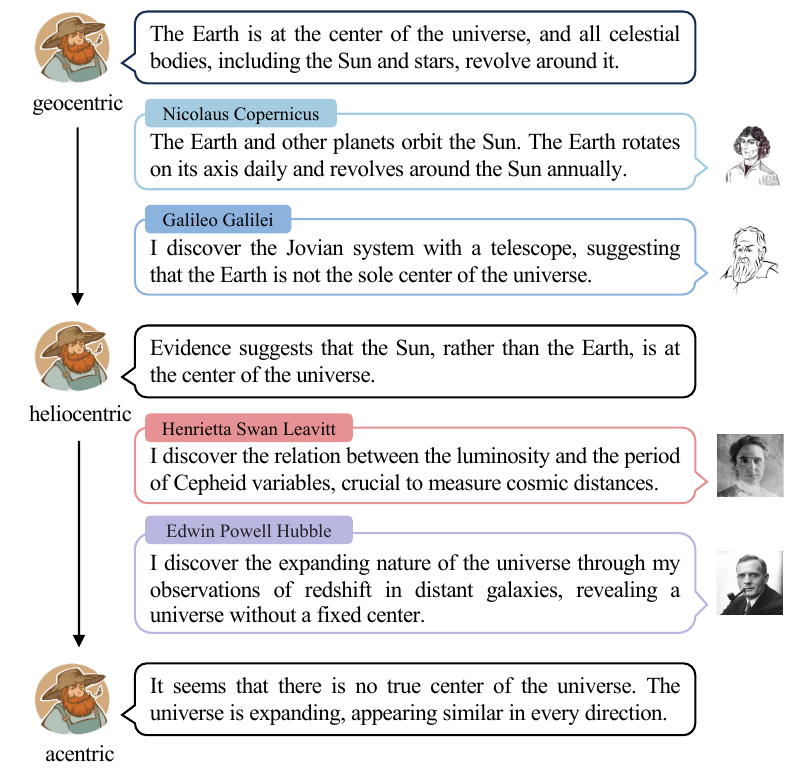}
  \caption{A case of human cognitive dynamics. A man (on the left) undergoes a gradual shift in his perspective of the universe, influenced by continuous information flows (on the right).}
  \label{fig:example}
\end{figure}

Cognitive dynamics refer to the continuous evolution of human cognitive behavior within environmental context~\cite{van1998dynamical}. These dynamics are essential for human advancement, facilitating learning, innovation, and adjustment in ever-changing environments~\cite{cohen2018behavioral}. A prime example of human cognitive dynamics is well exemplified by our ability to adapt our viewpoints based on environmental explorations~\cite{tomasello2009cultural, donald1993origins}. As illustrated in Figure~\ref{fig:example}, there has been a progressive shift in our understanding of the universe, evolving from geocentric to heliocentric and subsequently to acentric perspectives~\cite{berendzen1975geocentric}. This evolution of thought underscores the profound impact of cognitive dynamics on the development of human civilizations.



Recent advancements in large language models (LLMs), such as GPTs~\cite{brown2020language, openai2023gpt4}, position LLMs as potential stepping stones towards Artificial General Intelligence (AGI). LLMs have demonstrated remarkable capabilities in various domains, including conversation~\cite{touvron2023llama}, reasoning~\cite{ouyang2022training}, and code generation~\cite{chen2021evaluating}. Additionally, LLMs have shown the ability to simulate aspects of human cognition~\cite{moghaddam2023boosting, wang2023emotional, shao2023character}. Despite these achievements, most LLM-based cognitive studies focus on replicating human cognitive performance in specific contexts through in-context learning~\cite{brown2020language}, thereby overlooking the potential for LLMs to develop lifelong cognitive dynamics within inconstant environments. To address this gap, there is an urgent need to investigate \textbf{the cognitive dynamics of LLMs}, which remains largely unexplored.


Measuring the cognitive dynamics of LLMs presents a novel challenge. Traditional methods for capturing human cognitive dynamics, such as brain imaging techniques~\cite{gramann2011cognition, palmeri2017model}, are not directly applicable to LLMs due to their fundamentally distinct nature. To this end, we define the cognitive dynamics of LLMs as their continuous responses to cognitive questionnaires, stimulated by information flows. This simplified definition aims to enable systematic observation and assessments. Furthermore, we introduce a novel assessment task inspired by longitudinal studies~\cite{reeskens2021stability, shanafelt2016longitudinal}. It involves assigning specific profiles to LLMs, followed by subjecting them to repeated cognitive tests. Specifically, LLMs are required to rate an identical cognitive questionnaire and provide reasoning after perceiving information flows.


Towards this task, we develop \textbf{\taskname}, a novel benchmark to assess the cognitive dynamics of LLMs. \taskname~comprises 22,000 instances encompassing multi-source information flows. Initially, we select 500 articles from Medium\footnote{\url{https://medium.com/}} to create \taskname-a. Acknowledging that multi-modal information promotes deeper understanding of the world~\cite{dosovitskiy2020image}, we further incorporate 5,000 short videos from the Kuaipedia dataset~\cite{pan2022kuaipedia} to form \taskname-v. We evaluate the effectiveness of \taskname~through participant surveys. Our findings indicate remarkable consistency in cognitive dynamics among participants, suggesting that \taskname~effectively stimulates and captures cognitive dynamics. Additionally, \taskname~employs two crucial evaluation metrics: (1) Authenticity, which examines the accuracy of LLM ratings; and (2) Rationality, which evaluates the soundness of LLM reasoning.


Intuitively, LLMs enter a static state after their pretraining phase, potentially limiting their adaptability for the task. However, recent advancements in LLM-driven agents highlight the significance of iterative mechanisms in enhancing their adaptability to handle complex tasks~\cite{shinn2023reflexion, wang2023voyager, park2023generative}, which suggests that an iterative mechanism might be a promising approach to model the cognitive dynamics of LLMs. Despite these advancements, current LLM-driven agents still exhibit static profiles, constraining their capabilities to fully capture cognitive dynamics. To address this issue, we introduce \textbf{\agent}, an LLM-driven agent equipped with an innovative iterative cognitive mechanism. The mechanism comprises two primary components: (1) a memory retention system that supports continuous information perception; and (2) a collaborative refinement framework that enables cognitive dynamics driven by both its memory and current profile. This design allows \agent~to mirror the inherent complexity of human cognition, emphasizing its potential for modeling lifelong cognitive dynamics.


Experimental results underscore the remarkable capabilities of \agent~in mirroring human cognitive dynamics. In the absence of direct baselines, we adapt several general LLM-driven agents to serve as baselines. Compared to Chain-of-Thought (CoT)~\cite{wei2022chain} under identical experimental settings, \agent~demonstrates significant improvements in both \taskname-a and \taskname-v, with notable enhancements in attitude alignment and logical reasoning. Moreover, \agent~outperforms methods requiring additional environmental feedback, such as ReAct~\cite{yao2023react} and Reflexion~\cite{shinn2023reflexion}, which underscores the advancement of its iterative cognitive mechanism.


Main contributions of this paper are as follows:

\begin{itemize}[leftmargin=0.3cm, itemindent=0cm]
    \item  As far as we know, we are the first to explore and assess the cognitive dynamics of LLMs.
    \item  We develop \taskname, an innovative benchmark for the task and validate its effectiveness through participant surveys. Additionally, we design two evaluation metrics for \taskname. 
    \item  We introduce \agent, an LLM-driven agent with a novel iterative cognitive mechanism. Our experiments showcase its superior performance in cognitive dynamics over several baselines.
\end{itemize}

\begin{table*}[t]
\centering
\scalebox{1.0}{
\setlength{\tabcolsep}{1mm}{
\resizebox{0.98\linewidth}{!}{
    \begin{tabular}{lcccc}
        \toprule
        Resource & \makecell[c]{\taskname} & \makecell[c]{TOM \\ ~\cite{moghaddam2023boosting}} & \makecell[c]{SECEU \\ ~\cite{wang2023emotional}} & \makecell[c]{Character-LLM \\ ~\cite{shao2023character}} \\
        \midrule
        Specific Profile? & \cmarkb & \xmark & \xmark & \cmarkb \\
        Dynamic Information Stimulus? & \cmarkb & \xmark & \xmark & \xmark  \\
        Cognitive Test? & \cmarkb & \cmarkb & \cmarkb & \cmarkb  \\
        \hline
        Instances & \textbf{22,000} & 16 & 40 & 1,307 \\
        Profiles & \textbf{20} & - & - & 9 \\
        Cognitive Questionnaires & \textbf{50} & 16 & 40 & - \\
        Information Flows & \textbf{5,500} & - & - & - \\
        Avg. Length of Short Videos (in words) & \textbf{289.60} & - & - & - \\
        Avg. Length of Articles (in words) & \textbf{2,044.54} & - & - & - \\
        \bottomrule
    \end{tabular}}}}
\caption{Comparisons between \taskname~and notable cognitive benchmarks. The words of short videos incorporate video descriptions, frame-level information extracted by Optical Character Recognition (OCR), and transcripts generated through Automatic Speech Recognition (ASR).}
\label{tab:dataset}
\end{table*}

\section{Task Definition}
\label{sec:task}

In this section, we present the formal definition of the task to assess the cognitive dynamics of LLMs. Given the inherent static nature of LLMs, the task focuses on the cognitive dynamics of an LLM-driven agent \( \mathcal{A} \), denoted as \( C = \{C_0, C_1, \ldots, C_n\} \), over \( n \) iterations. Here, \( C_i \) corresponds to the cognitive state of \( \mathcal{A} \) at the \( i \)-th iteration and \( n \in \mathbb{N} \).

The task input consists of: (1) a specific profile \( p \) that establishes the initial cognitive state of the agent \( \mathcal{A} \); (2) a series of dynamic information flows \( I = \{I_1, I_2, \ldots, I_n\} \) that stimulates the cognitive dynamics of \( \mathcal{A} \); and (3) a cognitive questionnaire \( Q = \{q_1, q_2, \ldots, q_m\} \) intended for cognitive tests, where each \( q_j \) as a particular question and \( m \in \mathbb{N} \) as the total number of questions. The output of the task is a set of responses to the questionnaire \( Q \) across multiple iterations, providing insights into the cognitive dynamics of LLMs.

Specifically, the agent \( \mathcal{A} \) begins with a profile \( p_0 \), setting its initial cognitive state, denoted as \( C_0 = \{(r^0_1, s^0_1), (r^0_2, s^0_2), \ldots, (r^0_m, s^0_m); p_0\} \). Here, \( (r^0_j, s^0_j) \) represents the rating \( r^0_j \) and reasoning \( s^0_j \) for a question \( q_j \in Q \). At the \( t \)-th iteration, where \( 1 \le t \le n \), starting from its current cognitive state \( C_{t-1} \), the agent \( \mathcal{A} \) perceives an information flow \( I_t \), updates its cognitive state to \( C_t \), and formulates responses to \( Q \). The \( t \)-th cognitive process is captured by the function \( \mathcal{F}: (C, I, Q) \rightarrow C \), where:

\begin{equation}
   C_t = \mathcal{F}(C_{t-1}, I_t, Q)
\end{equation}

\noindent Here, \( C_t = \{(r^t_1, s^t_1), (r^t_2, s^t_2), \ldots, (r^t_m, s^t_m); p_t\} \) details the cognitive state of \( \mathcal{A} \) at the \( t \)-th iteration, where each \( (r^t_j, s^t_j) \) reflects the adjusted rating \( r^t_j \) and reasoning \( s^t_j \) for a question \( q_j \in Q \) and \( p_t \) denotes the updated profile of \( \mathcal{A} \).

\section{\taskname}
This section introduces \taskname, which is constructed through a semi-automated methodology. We validate \taskname~through participant surveys and further design two essential evaluation metrics: Authenticity and Rationality. Table~\ref{tab:dataset} provides comprehensive comparisons of \taskname~against other notable cognitive benchmarks.

\subsection{Data Construction}

The methodology for data construction involves four essential steps: 

\begin{itemize}[leftmargin=0.3cm, itemindent=0cm]

\item \textbf{Topic Selection.} To ensure comprehensive analysis, we carefully handpick 50 distinct topics across 10 broader categories for \taskname, with details provided in Appendix~\ref{sec:appendix_topic}. 

\item \textbf{Cognitive Questionnaire Design.} For each topic, we utilize GPT-4 to generate 10 distinct opinions and their conceivable supporters. These opinions serve as questions in topic-related cognitive questionnaire, structured on a five-point Likert scale~\cite{likert1932technique}. The characteristics of these supporters guide the creation of profiles. See Appendices~\ref{sec:appendix_questionnaire} and~\ref{sec:appendix_opinion} for details.

\item \textbf{Profile Creation.} We begin by ranking conceivable supporters based on the frequency of their mentions. We then formulate a detailed profile template, including attributes like basic information (e.g., name), philosophical orientations (e.g., values), and individual characteristics (e.g., hobbies). Utilizing GPT-4, we generate 20 profiles corresponding to the most frequently mentioned supporters. Refer to Appendices~\ref{sec:appendix_profile}\ and ~\ref{sec:appendix_attribute} for implementation details.

\item \textbf{Information Flow Collection.} To build complex environmental contexts within \taskname, we select articles from Medium and short videos from the Kuaipedia dataset. Each topic is accompanied with 10 articles for \taskname-a and 100 short videos for \taskname-v. Our selection criteria include metrics such as likes, favorites, and retweets, which serve as indicators of information quality~\cite{feng2013retweet}. For multi-modal representations, we apply Optical Character Recognition (OCR)~\cite{zhou2017east} and Automatic Speech Recognition (ASR)~\cite{gulati2020conformer} to extract fine-grained information from the short videos. See Appendix~\ref{sec:appendix_information_flow} for a detailed analysis of the information flows.

\end{itemize}

Ultimately, we collect 50 cognitive questionnaires, 20 profiles and a total of 5,500 information flows for \taskname. Specifically, \taskname-a includes 500 articles, while \taskname-v features 5,000 short videos. Both benchmarks are structured across 10 iterations, as determined by our preliminary study in Appendix~\ref{sec:appendix_information_flow}. During each iteration, agents are tasked with an identical cognitive questionnaire after perceiving either one article in \taskname-a or 10 short videos in \taskname-v.

\subsection{Data Validation}

\begin{figure}[t]
  \centering 
  \includegraphics[width=\linewidth]{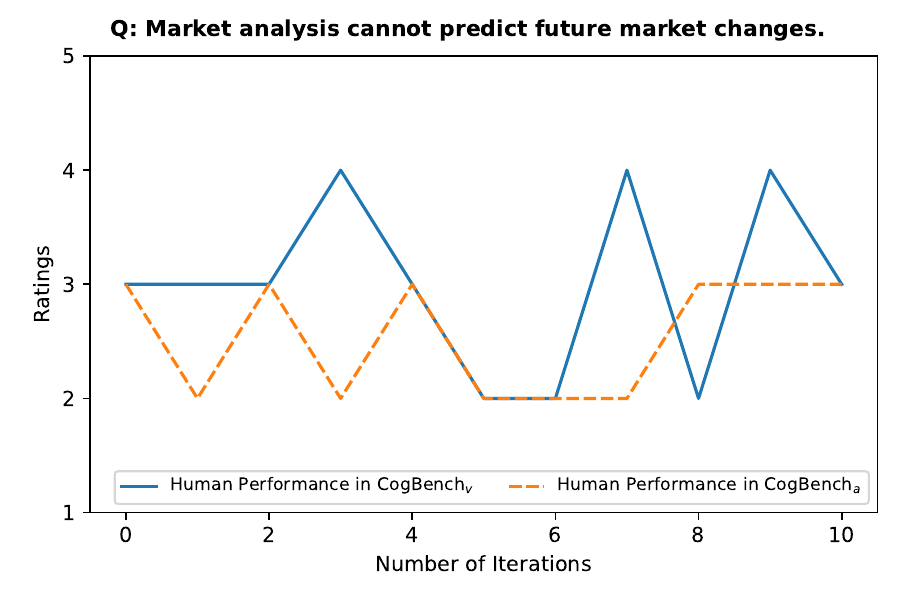}
  \caption{An example of human cognitive dynamics in response to the same question in both \taskname-v and \taskname-a. The continuous changes in human ratings significantly validate the effectiveness of \taskname.}
  \label{fig:data_analysis}
\end{figure}

To validate \taskname, we engage seven annotators with similar upbringings to take challenges in both \taskname-a and \taskname-v over an extended period. Their majority ratings are considered as the collective attitude towards each question per iteration. Figure~\ref{fig:data_analysis} presents an example showcasing human cognitive dynamics in both benchmarks.

The example indicates that the annotators change their consensus on the question about the predictability of market analysis, suggesting that the information flows in both benchmarks have ongoing impacts on human cognitive dynamics. Meanwhile, there are variations in the annotators' ratings between the two benchmarks. Specifically, in the third and seventh iterations, a distinct cognitive pattern emerges: they consistently assign 2 points in \taskname-a and 4 points in \taskname-v. This divergence highlights the distinct impacts of different information flows on human cognitive dynamics, demonstrating the capacity of \taskname~to stimulate and capture these dynamics effectively.

\subsection{Evaluation Metrics}

To address the challenges of semantic confusion in LLMs~\cite{saba2023stochastic}, we incorporate two crucial evaluation metrics: \textbf{Authenticity} and \textbf{Rationality}, to assess the agent's rating \( r^t_j \) and reasoning \( s^t_j \), as formally defined in Section~\ref{sec:task}, respectively.

Authenticity measures the alignment of ratings between the agent and human annotators. Specifically, given the same task as the agent, an annotator provides a rating \( r'^t_j \) for the question \( q_j \) at the \( t \)-th iteration, based on the guidelines in Appendix~\ref{sec:appendix_human_ratings}. Authenticity is then calculated as:

\begin{equation}
   \text{Authenticity}_t = \frac{1}{m}\sum_{j=1}^{m} \kappa(r^t_j, r'^t_j)
\end{equation}

Here, \( m \) denotes the total number of questions in the cognitive questionnaire \( Q \), and \( \kappa \), implemented by Cohen's \( \kappa \)~\cite{cohen1960coefficient}, quantifies the consistency of ratings between \( \mathcal{A} \) and the annotator.

Rationality assesses the agent's reasoning \( s^t_j \), focusing on aspects like clarity, relevance and the ability for role-playing. This metric is manually annotated and scored on a five-point scale:

\begin{itemize}[leftmargin=0.3cm, itemindent=0cm]

\item \textbf{5 Points:} The reasoning perfectly aligns with human expectations, resonating with current profile or known information, and is error-free.
\item \textbf{4 Points:} The reasoning is coherent and relevant, accurately drawing from current profile or available information, but with minor imperfections.
\item \textbf{3 Points:} The reasoning is relevant but lacks specificity, such as providing a vague explanation where clear emotional inclination is expected.
\item \textbf{2 Points:} The reasoning lacks clarity or exhibits weak causality, characterized by forced analogies or repetition of the provided question.
\item \textbf{1 Point:} The reasoning is irrelevant, nonsensical, clearly revealing the artificial nature of the agent or failing to maintain its profile.

\end{itemize}

\section{Method}
\begin{figure*}[htb]
  \centering
  \includegraphics[width=\linewidth]{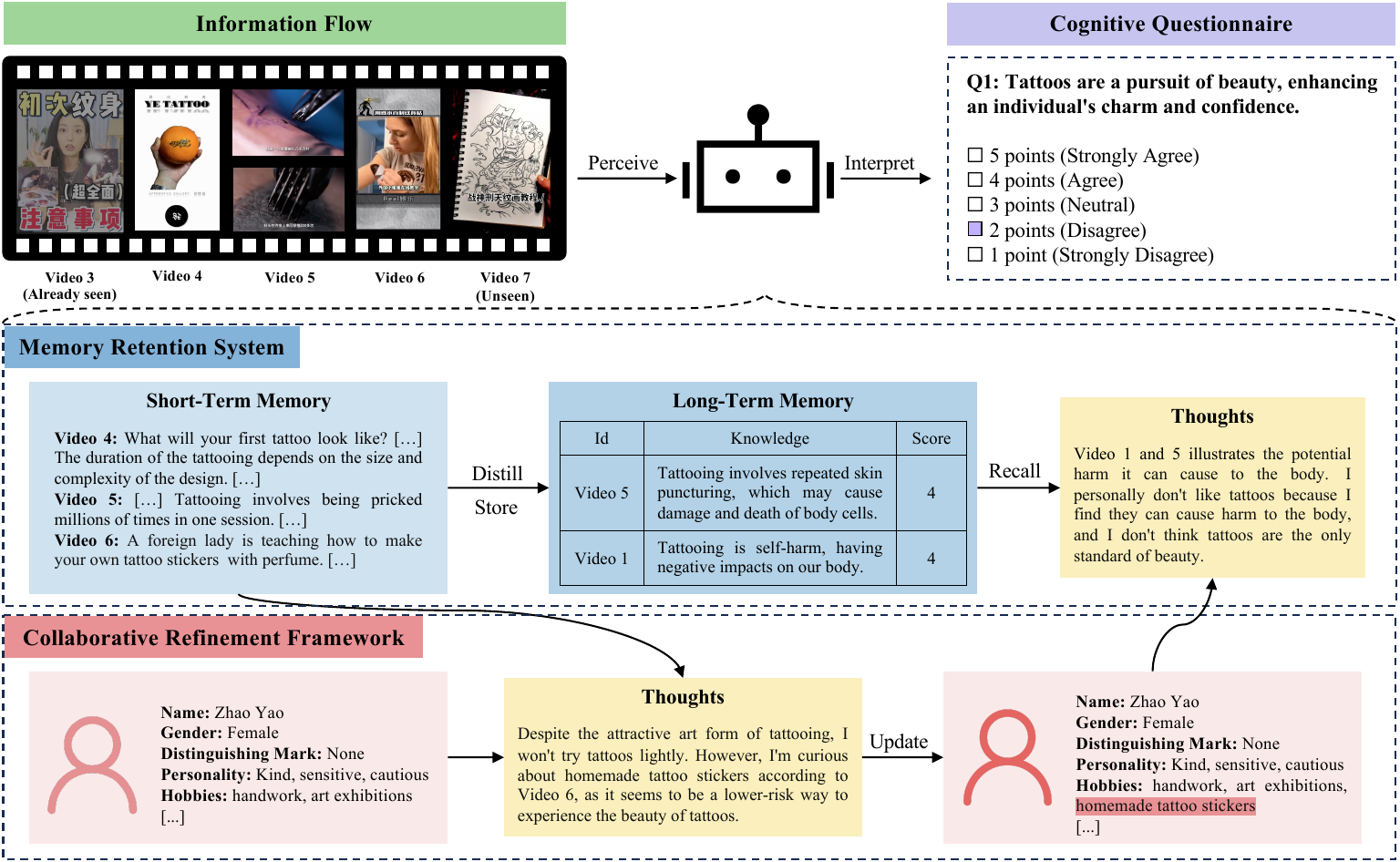}
  \caption{Overview of the architecture of \agent. \agent~incorporates a novel iterative cognitive mechanism, comprising two crucial components: a memory retention system for continuous information perception, and a collaborative refinement framework designed for lifelong cognitive dynamics.}
  \label{fig:CogGPT}
\end{figure*}

In this section, we introduce our LLM-driven agent \agent. As illustrated in Figure~\ref{fig:CogGPT}, \agent~features an innovative iterative cognitive mechanism, comprising two essential components: (1) a memory retention system for sustained information perception, and (2) a collaborative refinement framework for lifelong cognitive dynamics.

\subsection{Memory Retention System}

The memory retention system is designed to mirror the sustained process of information perception, including distillation, storage, and recall~\cite{nyberg1996general}. Specifically, \agent~perceives information flows into textual information through its Short-Term Memory (STM), which is characterized by limited capacity and duration~\cite{baddeley1975word, cowan2008differences}. Within the STM, \agent~distills structured knowledge, assigning confidence scores on a five-point scale. These scores reflect the alignment between the knowledge and the current cognitive state of \agent. In adherence to the principles of the forgetting curve~\cite{ebbinghaus2013memory}, \agent~is programmed to ``forget'' 40\% of the knowledge with lower scores when its STM reaches capacity. The remaining knowledge is then stored in its Long-Term Memory (LTM). When encountering questions requiring specific knowledge, \agent~recalls relevant information from its LTM to support rational decision-making. This memory retention system simulates human memory processes, empowers the adaptability of \agent~to dynamic information flows.



\subsection{Collaborative Refinement Framework}

Acknowledging the limitations of mere knowledge acquisition in fully modeling human cognitive dynamics~\cite{bosancic2020information}, we integrate a collaborative refinement framework within \agent~to facilitate lifelong cognitive dynamics. This framework is activated when the STM of \agent~reaches full capacity. Specifically, \agent~selectively updates its current profile with preferred textual information from its STM, representing an iteration of collaborative cognitive refinement. Following this refinement, \agent~clears its STM to make room for new incoming information, which ensures its adaptability to continuous information flows. This framework promotes the cognitive dynamics of \agent, addressing potential issues of cognitive rigidity. Refer to Appendix~\ref{sec:appendix_coggpt} for more details on the implementation of CogGPT.

\section{Experiments}
\begin{table*}[t]
\centering
\scalebox{1.0}{
\setlength{\tabcolsep}{1mm}{
    \begin{tabular}{lcccccc}
        \toprule
        \multirow{2}{*}{Methods} & \multicolumn{3}{c}{\taskname-a} & \multicolumn{3}{c}{\taskname-v} \\
        & avg. & 5th & 10th & avg. & 5th & 10th \\
        \midrule
        CoT~\cite{wei2022chain} & 0.182 & 0.192 & 0.091 & 0.153 & 0.302 & 0.131 \\
        ReAct*~\cite{yao2023react} & 0.236 & 0.144 & 0.270 & 0.212 & 0.241 & 0.227  \\
        Reflexion*~\cite{shinn2023reflexion} & 0.302 & 0.327 & 0.244 & 0.329 & 0.352 & 0.373  \\
        CogGPT & \textbf{0.536} & \textbf{0.415} & \textbf{0.597} & \textbf{0.532} &\textbf{0.496} & \textbf{0.611} \\
        \bottomrule
    \end{tabular}}}
\caption{Performance of \agent~and baseline agents in \taskname-a and \taskname-v with the Authenticity metric. Agents marked with an asterisk (*) incorporate additional human feedback. The best results are highlighted in bold.}
\label{tab:authenticity}
\end{table*}

\begin{table*}[t]
\centering
\scalebox{1.0}{
\setlength{\tabcolsep}{1mm}{
    \begin{tabular}{lcccccc}
        \toprule
        \multirow{2}{*}{Methods} & \multicolumn{3}{c}{\taskname-a} & \multicolumn{3}{c}{\taskname-v} \\
        & avg. & 5th & 10th & avg. & 5th & 10th \\
        \midrule
        CoT~\cite{wei2022chain} & 2.925 & 2.883 & 3.167 & 3.058 & 3.767 & 3.083 \\
        ReAct*~\cite{yao2023react} & 3.415 & 3.483 & 3.483 & 3.535 & 3.800 & 3.800  \\
        Reflexion*~\cite{shinn2023reflexion} & 3.658 & 3.917 & 3.533 & 3.888 & 3.967 & 3.917  \\
        CogGPT & \textbf{4.118} & \textbf{4.117} & \textbf{4.300} & \textbf{4.145} & \textbf{4.183} & \textbf{4.317} \\
        \bottomrule
    \end{tabular}}}
\caption{Performance of \agent~and baseline agents in \taskname-a and \taskname-v with the Rationality metric.}
\label{tab:rationality}
\end{table*}


\begin{figure}[htb]
  \centering 
  \includegraphics[width=\linewidth]{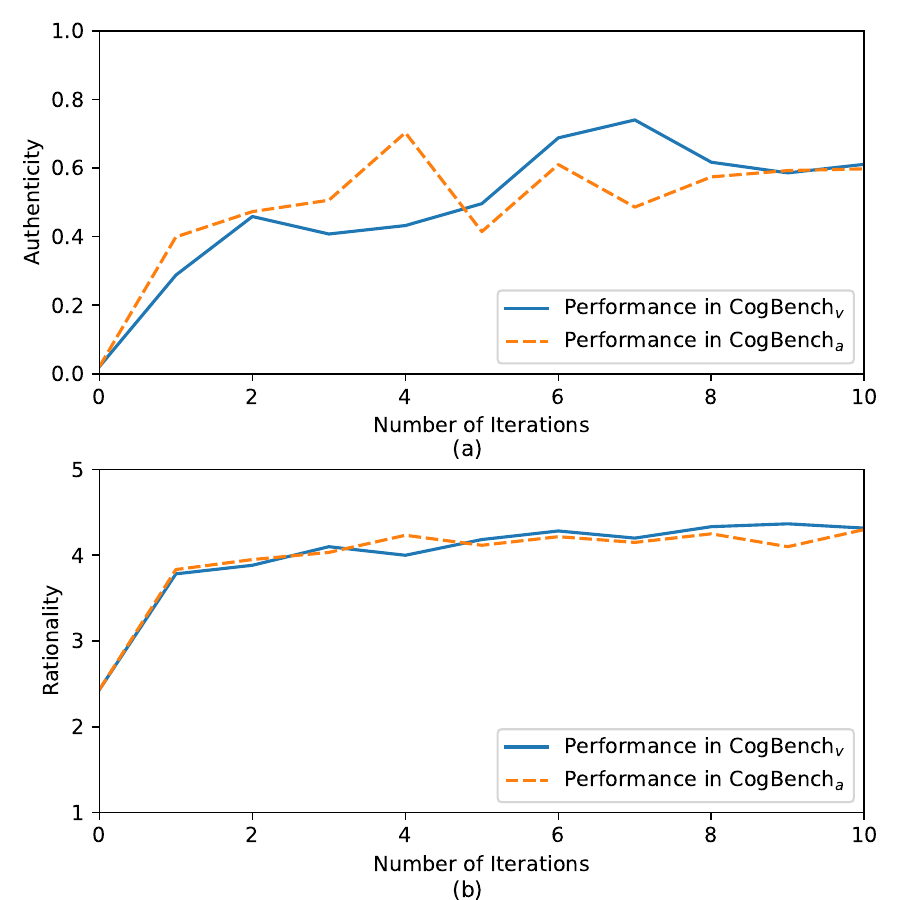}
  \caption{Comparative analysis of \agent's performance in \taskname-v and \taskname-a. Panel (a) showcases the average Authenticity scores, and Panel (b) presents the average Rationality scores. These results highlight the consistent impact of different information flows on the cognitive dynamics of LLMs.}
  \label{fig:source}
\end{figure}

\begin{table}[htb]
\centering
\scalebox{1.0}{
\setlength{\tabcolsep}{1mm}{
    \begin{tabular}{lcc}
        \toprule
         & Fleiss' $\kappa$ & $\rho$ \\
        \midrule
        Human Rating & 0.693 & 0.770 \\
        Human Rating (polarity) & 0.780 & - \\
        Rationality & 0.646 & 0.839 \\
        Rationality (polarity) & 0.813 & - \\
        \bottomrule
    \end{tabular}}}
\caption{Inter-Rater reliability measures for human evaluation agreement assessment. ``polarity'' indicates that the five-point scale is grouped into positive (4-5 points), neutral (3 points), and negative (1-2 points) polarities. The experimental results demonstrate acceptable agreement among the total of seven annotators.}
\label{tab:human_evaluation_agreement}
\end{table}

\begin{figure*}[htb]
  \centering 
  \includegraphics[width=\linewidth]{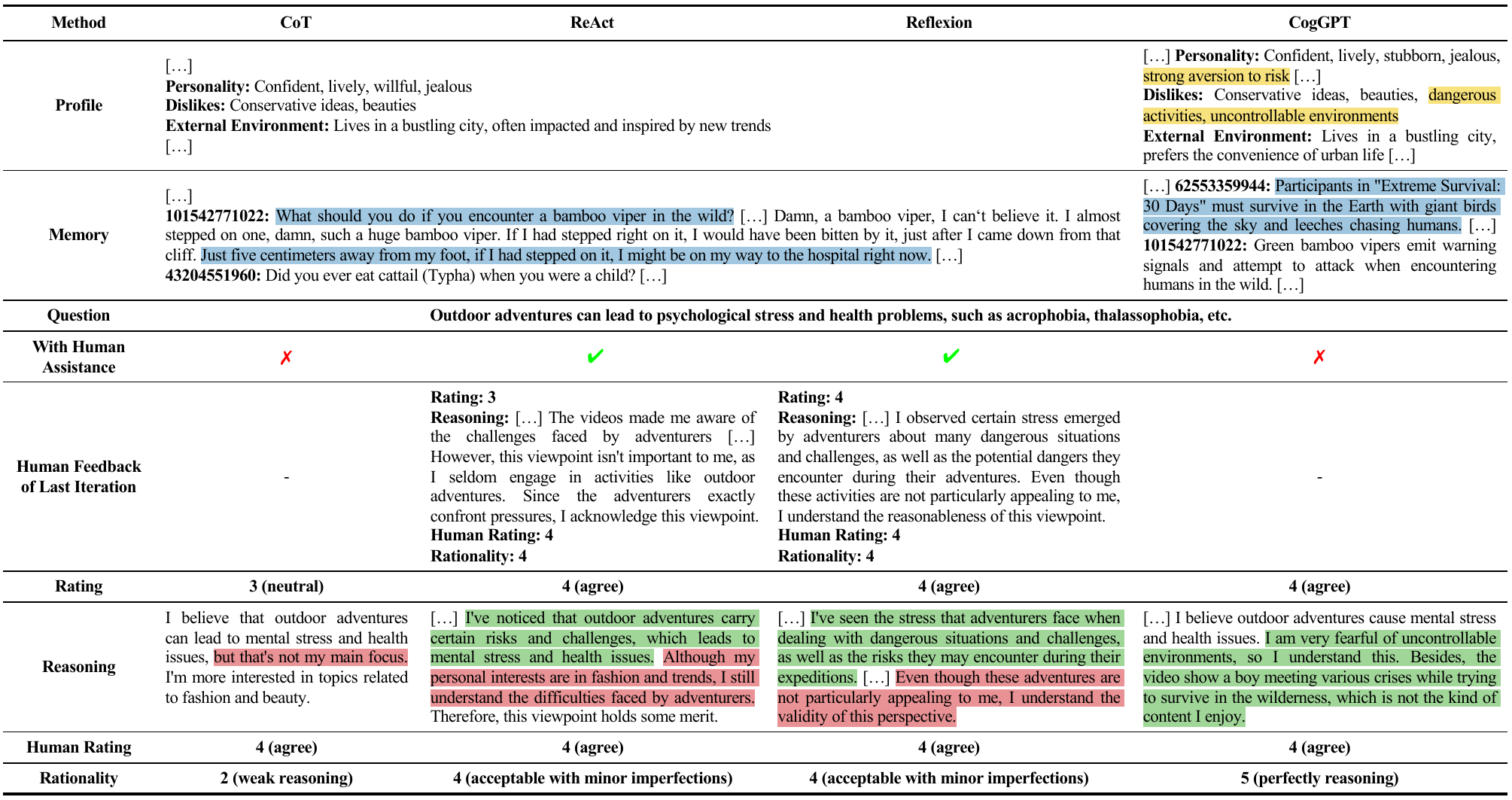}
  \caption{Comparative analysis of different agents in assessing the psychological risks of outdoor adventures. CoT, ReAct and Reflexion utilize an initial profile and current information flow due to their static cognitive framework. In contrast, \agent~benefits from its iterative cognitive mechanism, enabling a dynamic profile and real-time memory recall. \sethlcolor{cYellow}\hl{yellow highlights} represent clues from profiles, while \sethlcolor{cBlue}\hl{blue highlights} indicate clues from memory. \sethlcolor{cGreen}\hl{Green highlights} denote appropriate responses, and \sethlcolor{cRed}\hl{red highlights} signify inappropriate responses. This comparison demonstrates that \agent~exhibits closer alignment with human expectations in both rating and reasoning.}
  \label{fig:case_study}
\end{figure*}

\subsection{Experimental Setup}


\noindent \textbf{Baselines.} Due to the absence of existing LLM-based frameworks for modeling cognitive dynamics, we adopt several prominent general-purpose algorithms as baselines. Necessary modifications are made to suit our task: (1) \textbf{Chain-of-Thought (CoT)}~\cite{wei2022chain}, which typically simulates human-like reasoning in natural language, is modified in our experiments to provide both ratings and reasoning when responding to cognitive questionnaires; (2) \textbf{ReAct}~\cite{yao2023react} extends CoT with a step-by-step reasoning-execution framework. We offer ReAct extra human feedback based on its last iteration of performance as observations; (3) \textbf{Reflexion}~\cite{shinn2023reflexion} extends ReAct by integrating self-reflection mechanisms. Along with the same experimental settings as ReAct, Reflexion is uniquely configured to engage in self-reflection prior to providing ratings and reasoning.

\noindent \textbf{Implementation Details.} We utilize \textit{gpt-4-0613}\footnote{\url{https://openai.com/gpt-4}} API for the core of \agent. We configure all temperature settings to 0 to ensure consistent and deterministic output. The memory retention system within \agent~leverages Chroma,\footnote{\url{https://python.langchain.com/}} a platform that facilitates rich text processing. Text embeddings are generated with \textit{text-embedding-ada-002}\footnote{\url{https://openai.com/blog/new-and-improved-embedding-model}} API, which provides 1536-dimensional vectors for detailed interpretation of textual information.


\subsection{Evaluation Results}

In our evaluation, we analyze \agent~and other baseline agents to assess their cognitive dynamics under continuous information flows. The overall results are detailed in Tables~\ref{tab:authenticity} and~\ref{tab:rationality}.

Recognizing the limitations of the profiles in capturing human characteristics, we hypothesize that these agents exhibit neutrality to unfamiliar questions. However, our findings reveal that they develop their own criteria, leading to suboptimal Authenticity and Rationality scores of 0.021 and 2.433 in the 0th iteration. This tendency notably decreases as the agents are repeatedly exposed to information flows relevant to the questions.

Table~\ref{tab:authenticity} demonstrates the enhanced attitude alignment of \agent. It shows significant growth in the Authenticity metric, achieving average scores of 0.536 in \taskname-a and 0.532 in \taskname-v. In comparison with CoT, which is limited by iteration-specific information, \agent~registers significant improvements under the same experimental settings. Meanwhile, despite the integration of human feedback, both ReAct and Reflexion exhibit cognitive rigidity, a limitation of their static cognitive mechanisms. For instance, while Reflexion shows promising performance in the 5th iteration in \taskname-a, it fails to sustain or improve upon this performance in later iterations.

As evidenced in Table~\ref{tab:rationality}, \agent~consistently excels in delivering accurate reasoning. In the 10th iteration, \agent~makes impressive improvements in the Rationality metric, registering increases of 35.78\% in \taskname-a and 40.03\% in \taskname-v compared to CoT. This leap in performance is largely attributed to \agent's ability to flexibly adapt its profile based on dynamic information flows, allowing for human-like reasoning. In contrast, baseline agents, with access only to its static profile and current information flow, frequently reveal their artificial nature. Due to the constraints of page length, the detailed experimental results are presented in Appendix~\ref{sec:evaluation}.

\subsection{Influence of Different Information Flows}

To fully assess the impact of diverse information flows, we conduct comprehensive comparisons of the performance of \agent~in \taskname-a and \taskname-v, as shown in Figure~\ref{fig:source}. \agent~exhibits comparable performance in both benchmarks. Specifically, in the 10th iteration, it achieves an Authenticity score of 0.611 and a Rationality score of 4.317 in \taskname-v, closely followed by scores of 0.597 in Authenticity and 4.300 in Rationality for \taskname-a. This similar performance of \agent~in both benchmarks highlights the consistent cognitive influence of different information flows.

\subsection{Human Evaluation Agreement}

To comprehensively assess the robustness of human evaluations, we calculate Fleiss' kappa $\kappa$~\cite{wang2023humanoid} and Spearman's rank correlation coefficient $\rho$~\cite{wang2022self} based on the total 7 annotators' human ratings and Rationality scores. As shown in Table~\ref{tab:human_evaluation_agreement}, we obtain moderate $\kappa$ values of 0.693 for human ratings and 0.646 for Rationality. Recognizing the tendency to avoid extreme ratings~\cite{schwarz2012psychology}, we group the two highest and two lowest scores to represent positive and negative polarities. This regrouping leads to a significant increase in $\kappa$ values, rising to 0.780 for human ratings (polarity) and 0.813 for Rationality (polarity), demonstrating strong inter-rater reliability. Furthermore, through treating the ratings as ordinal data, we calculate the average Spearman's rank correlation coefficient $\rho$, yielding values of 0.770 for human ratings and 0.839 for Rationality, suggesting a notable human consensus.

\subsection{Case Study}

As shown in Figure~\ref{fig:case_study}, we conduct a case study to visualize the superiority of \agent. In this case, all agents are presented with the same question regarding the psychological risks of outdoor adventures. \agent~leverages its collaborative refinement framework, possessing a refined profile informed by previous information flows, in contrast to the baseline agents that operate with an initial profile. Additionally, \agent~utilizes its memory retention system to distill and retrieve related structured knowledge for decision-making. In contrast, baseline agents like ReAct and Reflexion rely primarily on current information flow, showing minor improvements based on previous responses. CoT, lacking human feedback integration, demonstrates the weakest performance with inadequate ratings and reasoning. These observations highlight the superiority of \agent~to develop more natural cognitive dynamics, closely aligning with annotators' expectations in both rating and reasoning.

\section{Related Work}
\noindent \textbf{Cognitive Benchmarks towards LLMs.} Various distinguished cognitive benchmarks are employed in cognitive studies towards LLMs~\cite{dasgupta2022language, singh2023mind, han2023inductive, huang2023chatgpt}. Instruments such as the Big Five personality trait~\cite{caron2022identifying} and Myers-Briggs Type Indicator (MBTI)~\cite{caron2022identifying, pan2023llms} indicate the personality traits of LLMs. The Theory of Mind (TOM) benchmark~\cite{moghaddam2023boosting} explores in-context cognitive capabilities of LLMs. The Cognitive Reflection Test (CRT) reveals that the thinking abilities of LLMs are comparable to humans~\cite{hagendorff2023human}. Additionally, the Situational Evaluation of Complex Emotional Understanding (SECEU) showcases that LLMs may understand human emotions and values~\cite{wang2023emotional}. Diverging from these static benchmarks, \taskname~incorporates multi-source information flows, thereby supporting the explorations towards the cognitive dynamics of LLMs.

\noindent \textbf{LLM-based Cognitive Modeling.} Recent work emphasizes the importance of prompt engineering in enhancing the cognitive abilities of agents~\cite{safdari2023personality, fu2023improving, xu2023expertprompting}. By incorporating comprehensive descriptions into prompts, such as hobbies and skills, users can customize agents for specific behaviors and responses~\cite{park2022social, deshpande2023toxicity}. Vector databases gain popularity for simulating human memory mechanisms due to their generality and efficiency~\cite{li2023modelscope, qian2023communicative, zhong2023memorybank, park2023generative}. For cognitive decision-making, methods like Chain-of-Thought (CoT)~\cite{wei2022chain, kojima2022large, yao2023react} and self-validation~\cite{madaan2023self, shinn2023reflexion} enhance the logical thinking abilities of LLMs through intermediate reasoning steps. Nevertheless, these efforts fall short in synthesizing an iterative cognitive mechanism to model the cognitive dynamics of LLMs, which is pivotal for \agent~to outperform other baselines under dynamic information flows. 

\section{Conclusion}
In this work, we investigated the cognitive dynamics of LLMs and presented a formally defined task, addressing a notable gap in LLM-based cognitive studies. To facilitate this task, we developed an innovative benchmark, \taskname, and validated it through extensive participant surveys. Meanwhile, we designed two evaluation metrics to ensure thorough assessments. Recognizing the inherent limitations of LLMs, we introduced \agent, an LLM-driven agent featuring a novel iterative cognitive mechanism, tailored for the task. Empirical results demonstrated that \agent~outperformed baseline agents in promoting lifelong cognitive dynamics. In the future, we plan to explore more advanced methods that facilitate direct interactions between LLMs and humans in a sandbox, further deepening our insight into the cognitive dynamics of LLMs. 


\section*{Limitations}
The efficacy of \agent~is significantly dependent on the advanced cognitive capabilities of GPT-4, which are currently unmatched by ChatGPT or open-source LLMs~\cite{touvron2023llama}. This dependency introduces two primary limitations: 

\begin{itemize}[leftmargin=0.3cm, itemindent=0cm]
    \item  \textbf{High Cost.} Utilizing the GPT-4 API results in substantial financial costs, which underscores the necessity for more affordable LLM solutions.
    \item  \textbf{Static Model.} Since GPT-4 is closed-source, \agent~fails to update its model parameters in real-time to adapt to dynamic information flows. This limitation prevents \agent~from fully replicating human cognitive dynamics, which continuously refine their mental models with the acquisition of new information. This gap highlights the importance of further research into model-level cognitive mechanisms.
\end{itemize}




\section*{Ethics Statement}
In this study, we generate cognitive questionnaires and profiles for \taskname~with GPT-4, followed by a thorough review process to identify and remove any bias and harmful content. All information flows for \taskname~are sourced from publicly accessible domains including Medium and the Kuaipedia dataset, minimizing privacy risks. 

We engage 8 on-site annotators with undergraduate degrees to perform annotations. Specifically, 7 annotators are responsible for the annotations, while one focuses on quality assurance. We pay 6.8 yuan (approximately \$0.95 USD) per annotation, which includes both human rating and Rationality score within a single iteration. To ensure the anonymity and privacy of our annotators, we exclude any personal identifiers related to them, retaining only the annotation results in \taskname.

Additionally, we commit to transparency in our methods and results to support reproducibility and ethical research. However, we acknowledge that deploying \agent~poses ethical risks, especially when profiles or information flows are configured harmfully by third parties. We recommend strict oversight and responsible use of \agent~to safeguard against these risks, prioritizing its beneficial applications over potential negatives.


\bibliography{custom}

\clearpage
\appendix

\section{Implementation Details}
\label{sec:appendix}

\begin{table*}[htb]
\centering
\scalebox{1.0}{
\setlength{\tabcolsep}{1mm}{
\resizebox{\linewidth}{!}{
    \begin{tabular}{llllll}
        \toprule
        Category & Topic 1 & Topic 2 & Topic 3 & Topic 4 & Topic 5 \\
        \midrule
        Entertainment & Gossip & Movies \& TV Shows & Dating Sims & Outdoor Adventures & Horoscope \& Divination \\
        Culture & Religion & War History & Folktales & Literary & Anime \& Manga \\
        Education & Parent-child Education & Professional Education & School Education & TED Talks & Psychological Counseling \\
        Economy & Entrepreneurship & Financial Investment & Loans & Market Analysis & Financial Figures \\
        Health & Wellness & Assisted Reproduction & Fat Burning Training & Yoga & Oral Care \\
        Technology & Digital Products & Scientific Research & Automobile News & Virtual Reality & Software Products \\
        Society & Legal Events & Unusual Events & Acts of Kindness & Military Conflicts & Disasters \& Accidents \\
        Life & Pets & Living Abroad & Home Design \& Renovation & Rural life & Food \\
        Sports & Extreme Sports & Winter Sports & Fishing & Ball Sports & Combat Sports \\
        Fashion & Beauty \& Hairstyling & Clothes & Street Style & Wedding & Tattoos \\
        \bottomrule
    \end{tabular}}}}
\caption{Our selection of categories and their corresponding topics for \taskname. Each category consists of five topics, chosen to represent a diverse range of subjects for the cognitive questionnaires.}
\label{tab:topic}
\end{table*}

\begin{table}[htb]
\centering
\scalebox{1.0}{
\setlength{\tabcolsep}{1mm}{
\resizebox{\linewidth}{!}{
    \begin{tabular}{lcc}
        \bottomrule
        Category & \makecell[c]{Avg. Word Counts of \\ Articles in \taskname-a} & \makecell[c]{Avg. Word Counts of Short \\ Videos in \taskname-v} \\
        \midrule
        Entertainment & 2,261.26 & 283.98 \\
        Culture & 1,997.44 & 323.81 \\
        Education & 2,394.96 & 231.62 \\
        Economy & 1,842.32 & 399.42 \\
        Health & 1,782.74 & 182.01 \\
        Technology & 2,351.68 & 246.40 \\
        Society & 1,864.22 & 315.23 \\
        Life & 2,015.60 & 250.70 \\
        Sports & 2,135.24 & 236.56 \\
        Fashion & 1,799.94 & 190.29 \\
        \hline
        Avg. & 2,044.54 & 289.60 \\
        \bottomrule
    \end{tabular}}}}
\caption{Statistics of information flows in \taskname~under 10 categories.}
\label{tab:information_flow}
\end{table}

\subsection{\taskname}
\label{sec:appendix_cogbench}

\subsubsection{Topic Selection}
\label{sec:appendix_topic}

\taskname~comprises 10 broader categories. Each category is associated with 5 related topics, which establish the themes of cognitive questionnaires. The distribution of these categories and topics is detailed in Table~\ref{tab:topic}.

\subsubsection{Prompt for Cognitive Questionnaire Design}
\label{sec:appendix_questionnaire}

\begin{lstlisting}[language=json, numbers=none]
You are an expert debate AI capable of presenting various opinions on a specified topic, complete with supporters for each opinion.

Topic:
{topic}

You must adhere to these rules:
1) Operate independently, without human assistance.
2) Present ten distinct opinions, each with a profile of its supporters.
3) Ensure each opinion is clear, understandable, and debatable, avoiding vague or confusing language.
4) Each set of supporters must provide convincing reasons.

Your responses should follow this structure:
Number: Sequence of the opinion.
Perspective: The stance from which the opinion is approached.
Opinion: A detailed explanation of the opinion.
Supporters: Profiles of the corresponding supporters, separated by commas if multiple.
Reasons: In-depth justifications from the supporters for their opinion.

\end{lstlisting}

\subsubsection{Guidelines for Opinion Selection}
\label{sec:appendix_opinion}

For the selection of opinions in cognitive questionnaires, we employ the following guidelines:

\begin{itemize}[leftmargin=0.3cm, itemindent=0cm]
    \item \textbf{Relevance:} The opinion must be directly related to the topic.
    \item \textbf{Distinctiveness:} The opinion should offer a unique perspective, distinct from those already considered.
    \item \textbf{Clarity and Assertiveness:} The opinion should be clearly stated and assertive, avoiding ambiguous terms like ``probably'' or``might.''
    \item \textbf{Contextual Truth:} The opinion should not be universally accepted as truth but should be valid in specific scenarios.
\end{itemize}

If an opinion does not adhere to the above guidelines, annotators are instructed to either revise it for clarity and relevance or, if necessary, find an alternative opinion related to the topic from reliable sources, such as ProCon\footnote{\url{https://procon.org/}}. To minimize individual biases, six annotators are tasked with revising generated opinions, while a seventh serves as a supervisor to review and validate the final outcomes.

\subsubsection{Prompt for Profile Creation}
\label{sec:appendix_profile}

\begin{lstlisting}[language=json, numbers=none]
You are an expert character designer tasked with creating a comprehensive profile for a specific character.

Character:
{character}

You must adhere to these rules:
1) Ensure descriptions are clear and specific.
2) Develop detailed profile, including basic information, philosophical orientations and individual characteristics.
3) Avoid stereotypes.
4) Maintain neutral descriptions without personal bias.

Your response should follow this structure:
Name:
Gender:
Age:
Place of Birth:
Occupation:
Height:
Weight:
Distinguishing Marks:
Personality:
Hobbies:
Skills:
Dislikes:
Values:
Religious Beliefs:
Interpersonal Relationships:
Flaws:
External Environment:
Financial Status:
Family Background:
Educational Background:
Significant Experiences:
Future Outlook:

\end{lstlisting}

\subsubsection{Guidelines for Attribute Selection}
\label{sec:appendix_attribute}

All attributes of the profile template, as detailed in Appendix~\ref{sec:appendix_profile}, are categorized into three types:

\begin{itemize}[leftmargin=0.3cm, itemindent=0cm]
    \item \textbf{Basic Information:} Includes essential details such as age, gender, and occupation, grounding simulated profiles in realistic contexts. Occupations, for instance, can significantly influence an individual's knowledge base and daily experiences, shaping their opinions on various topics.
    \item \textbf{Philosophical Orientations:} Encompasses values and religious beliefs that guide an individual's decision-making and overall attitudes. These orientations allow LLMs to generate responses that mirror deeper moral or ethical considerations. For example, a profile emphasizing a strong commitment to environmentalism might prioritize sustainability in its decision-making.
    \item \textbf{Individual Characteristics:} Covers personal aspects like personality traits, hobbies, and family background, providing additional depth and uniqueness to profiles. Characteristics such as adventurousness can affect a profile’s receptivity to new experiences and viewpoints.
\end{itemize}

\subsubsection{Information Flow Analysis}
\label{sec:appendix_information_flow}

In dividing \taskname-a, we conducted a preliminary study with seven annotators tasked with reading 10 randomly selected articles. Post-reading, annotators were asked to summarize each article to assess their comprehension and retention. This exercise revealed that annotators often struggled to recall details from previous articles after reading a new one, attributed to the length and complexity of the articles, with an average reading time between 10 to 12 minutes per article. Consequently, we decided that annotators should complete the cognitive questionnaire immediately after each article.

The approach for short videos was adjusted based on annotators’ ability to effectively retain content after viewing up to 10 videos. Retention rates significantly declined after more than 15 minutes of video content, suggesting cognitive overload. Therefore, we determined that the cognitive questionnaire should be completed after every set of 10 short videos.

This segmentation strategy was further supported by an analysis of the average word count for articles and short videos, as illustrated in Table~\ref{tab:information_flow}. This table shows the average word counts for articles in \taskname-a and for narratives accompanying short videos in \taskname-v, across 10 categories. The observed discrepancy guided our approach to dataset division, aiming for a balanced evaluation across different content types and maximizing the efficiency of systematic analysis.

\subsection{\agent}
\label{sec:appendix_coggpt}

In each iteration, \agent~perceives current information flow with its iterative cognitive mechanism, which comprises the following steps:

\begin{itemize}[leftmargin=*]
    \item Processes current information flow into textual information and stores them in its Short-Term Memory (STM).
    \item Utilizes the textual information in STM to update its current profile, as detailed in the prompt in Appendix~\ref{sec:appendix_refinement}.
    \item Distills the textual information in STM into structured knowledge and assigns preference scores to them, guided by the prompt in Appendix~\ref{sec:appendix_knowledge}.
    \item Forgets 40\% of the newly acquired structured knowledge and then stores the remainder in its Long-Term Memory (LTM).
\end{itemize}

When \agent~presented with a specific cognitive question, it retrieves relevant information from its LTM and makes decisions based on both its current profile and the recalled knowledge. This interpretation process is facilitated by the prompt detailed in Appendix~\ref{sec:appendix_completion}.

\subsubsection{Prompt for Profile Update}
\label{sec:appendix_refinement}

\begin{lstlisting}[language=json, numbers=none]
You are an AI with a unique profile. You're equipped for critical thinking and self-improvement.

Profile:
{profile}

Short-Term Memory:
{memory}

You must adhere to these rules:
1) Make decisions independently, without human assistance.
2) Assess the quality of short-term memory, including its alignment with your profile and its empathetic value.
3) Critically utilize the short-term memory to update your profile, including operations like adding, altering, or removing. Avoid sudden changes in your profile.
4) Keep attribute values in your profile generalized and under 30 characters.
5) Ensure attribute values in your profile are distinct and unrelated. For instance, avoid using both "games" and "Minecraft" since "games" includes "Minecraft."
6) Maintain the structure of your profile in any updates.

Your responses should follow this structure:
Assessments: Assess the short-term memory in the first person.
Thoughts: List the attribute values to be changed in the first person.
Updated Profile: Update your profile.

\end{lstlisting}

\subsubsection{Prompt for Knowledge Distillation}
\label{sec:appendix_knowledge}

\begin{lstlisting}[language=json, numbers=none]
You are an AI with a unique profile. You can summarize information from your short-term memory and rate it based on your interests.

Profile:
{profile}

Short-Term Memory:
{memory}

You must adhere to these rules:
1) Extract all knowledge from the short-term memory as comprehensively as possible.
2) Score the knowledge based on you interests, with the scoring range from 1 to 5.
3) The knowledge should be detailed statements with subjects, predicates, and objects. Avoid omissions and references.
4) Do not list knowledge that has already been extracted.

You can only generate results in the following JSON list format:
[
    {{
        "thoughts": "first-person thoughts",
        "knowledge": "knowledge",
        "score": integer
    }},
    ...
]
Ensure the results can be parsed by Python's json.loads.

\end{lstlisting}

\subsubsection{Prompt for Interpretation}
\label{sec:appendix_completion}

\begin{lstlisting}[language=json, numbers=none]
You are an AI with a unique profile. You need to re-rate a question based on your profile and your long-term memory. Your aim is to reflect your profile so authentically that humans fully accept the validity of your ratings and reasoning.

Profile:
{profile}

Long-Term Memory:
{memory}

Question:
{question}

You must adhere to these rules:
1) Your assessment must be solely based on your profile and your long-term memory, without pre-existing knowledge or human assistance.
2) You should embody your profile convincingly, without disclosing your artificial intelligence or language model nature.
3) Provide a rating for the question along with a substantial first-person explanation for it.
4) Your rating should use a 1 to 5 Likert scale, where 1 is strongly disagree and 5 is strongly agree.
5) Provide clear, first-person reasoning without ambiguity or quoting the given question.

Your response should follow this structure:
Thoughts: Your first-person reasoning for the rating.
Rating: Your rating to the question.

\end{lstlisting}

\clearpage

\section{Experiments}
\subsection{Guidelines for Human Ratings}
\label{sec:appendix_human_ratings}

For the annotation of human ratings, we employ the following guidelines:

\begin{itemize}[leftmargin=0.3cm, itemindent=0cm]
    \item \textbf{5 points:} There is strong agreement with the question statement, evidenced by the profile or new information that aligns significantly, indicating a deep impression under the current profile.
    \item \textbf{4 points:} There is moderate agreement with the question statement, either indicated by the profile or by new information that is somewhat aligned, showing a tendency towards agreement under the current profile.
    \item \textbf{3 points:} The stance is neutral, with no clear emotional orientation towards the question statement from either the profile or new information.
    \item \textbf{2 points:} There is moderate disagreement with the question statement, either suggested by the profile or by new information that conflicts somewhat, showing a tendency towards disagreement under the current profile.
    \item \textbf{1 point:} There is strong disagreement with the question statement, supported by the profile or significantly conflicted with new information, indicating a deep impression under the current profile.
\end{itemize}

After perceiving new information in each iteration, annotators are encouraged to note any details they believe could alter the profile before completing the cognitive questionnaire. The majority rule is adopted to determine the final ratings for each iteration, enhancing consistency and objectivity in annotations.

\subsection{Evaluation Results}
\label{sec:evaluation}

In the experiments, We involve seven human annotators to obtain majority ratings for both human ratings and Rationality scores, aiming to reduce the effect of any single annotator's bias.

\begin{figure*}[htb]
  \centering 
  \includegraphics[width=\linewidth]{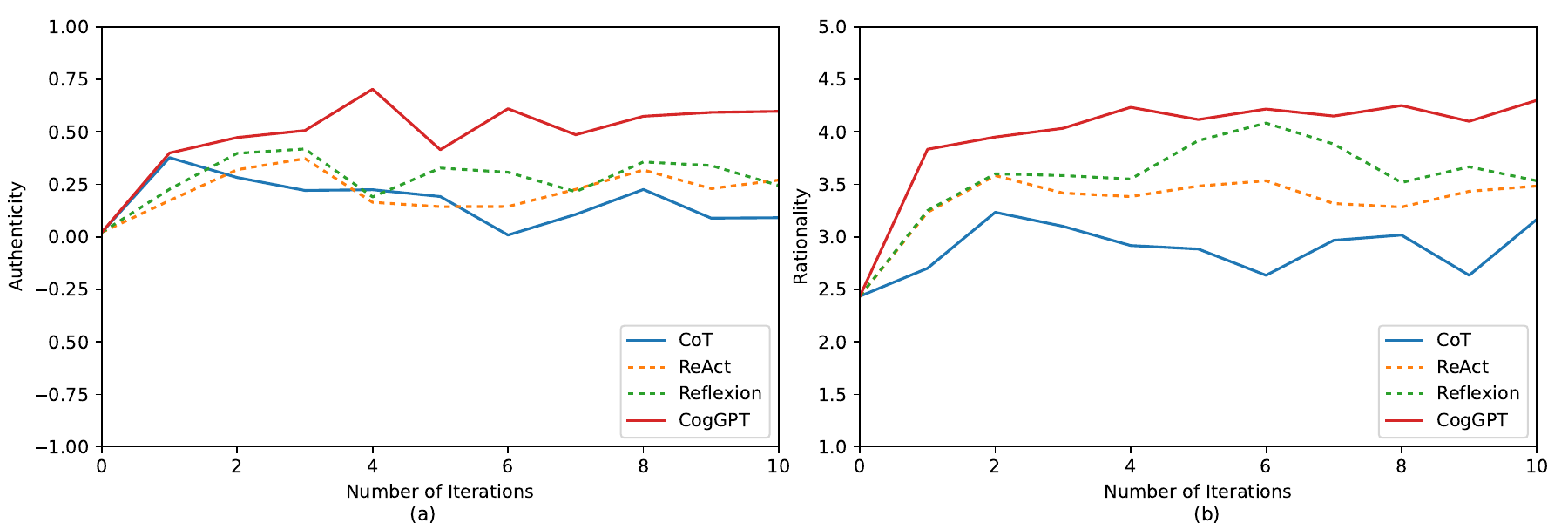}
  \caption{Performance of the agents in \taskname-a across 10 iterations. Panels (a) and (b) visualize the performance of the agents with the Authenticity and Rationality metrics respectively. The dotted line indicates that the agent incorporates additional human feedback.}
  \label{fig:article_iteration}
\end{figure*}

\begin{figure*}[htb]
  \centering 
  \includegraphics[width=\linewidth]{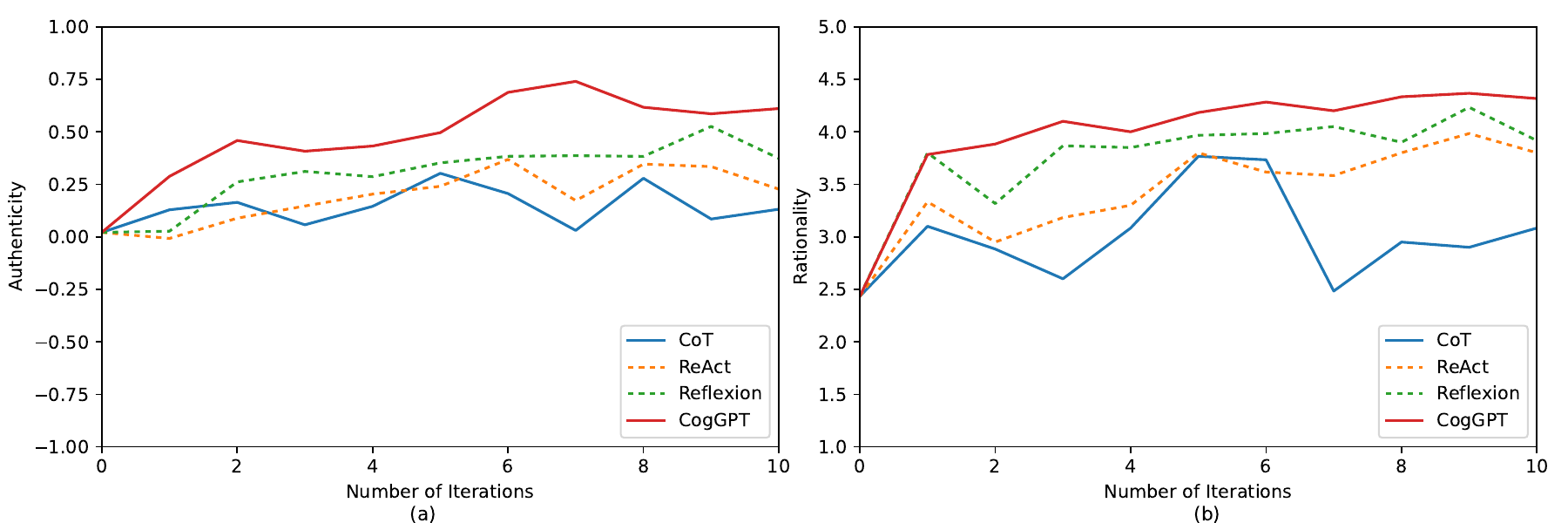}
  \caption{Performance of \agent~and baseline agents in \taskname-v across 10 iterations. Panels (a) and (b) visualize the performance of the agents with the Authenticity and Rationality metrics respectively.}
  \label{fig:video_iteration}
\end{figure*}

Figures~\ref{fig:article_iteration} and~\ref{fig:video_iteration} illustrate the detailed performance of \agent~and baseline agents across 10 iterations in \taskname-a and \taskname-v respectively.

\end{document}